\documentclass[letterpaper]{article} 
\usepackage{aaai2026}  
\usepackage{times}  
\usepackage{helvet}  
\usepackage{courier}  
\usepackage[hyphens]{url}  
\usepackage{graphicx} 
\usepackage{natbib}  
\usepackage{caption} 
\usepackage{algorithm}
\usepackage{algorithmic}

\usepackage{amsfonts}
\usepackage{amsmath}
\usepackage{booktabs}

\usepackage{xcolor}

\usepackage{newfloat}
\usepackage{listings}
\DeclareCaptionStyle{ruled}{labelfont=normalfont,labelsep=colon,strut=off} 
\floatstyle{ruled}
\newfloat{listing}{tb}{lst}{}
\floatname{listing}{Listing}

\title{E-React: Towards Emotionally Controlled Synthesis of Human Reactions}
\author{
    Chen Zhu\equalcontrib,
    Buzhen Huang\equalcontrib,
    Zijing Wu,
    Binghui Zuo,
    Yangang Wang\thanks{Corresponding authors}
}
\affiliations{
    Southeast University, China\\
    chenzhu@seu.edu.cn, buzhenhuang@outlook.com, zijing55555@gmail.com, binghuizuo@gmail.com, yangangwang@seu.edu.cn
}

\begin{document}

\maketitle

\begin{abstract}
Emotion serves as an essential component in daily human interactions. Existing human motion generation frameworks do not consider the impact of emotions, which reduces naturalness and limits their application in interactive tasks, such as human reaction synthesis. In this work, we introduce a novel task: generating diverse reaction motions in response to different emotional cues. However, learning emotion representation from limited motion data and incorporating it into a motion generation framework remains a challenging problem. To address the above obstacles, we introduce a semi-supervised emotion prior in an actor-reactor diffusion model to facilitate emotion-driven reaction synthesis. Specifically, based on the observation that motion clips within a short sequence tend to share the same emotion, we first devise a semi-supervised learning framework to train an emotion prior. With this prior, we further train an actor-reactor diffusion model to generate reactions by considering both spatial interaction and emotional response. Finally, given a motion sequence of an actor, our approach can generate realistic reactions under various emotional conditions. Experimental results demonstrate that our model outperforms existing reaction generation methods. The code and data will be made publicly available at \url{https://ereact.github.io/}
\end{abstract}

\section{Introduction}
\label{sec:introduction}

Human-to-human interaction perception and generation play a crucial role in the development of future digital humans. During social interactions, emotions significantly influence interaction dynamics~\cite{james2004principles, yoo2006influence, boutet2021emojis, dukes2021rise} and may lead to totally different motions. However, most existing motion generation works~\cite{tevet2023human, xu2024inter, karunratanakul2023guided, zhu2023taming, kim2022brand, zhang2024motiongpt} neglect the emotional dynamics and often result in rigid and emotion-agnostic motions. Therefore, modeling interaction patterns under diverse emotional states is a crucial yet under-explored direction for enhancing the naturalness and expressiveness of multi-person motion generation.

\begin{figure}
  \centering
  \includegraphics[width=\linewidth]{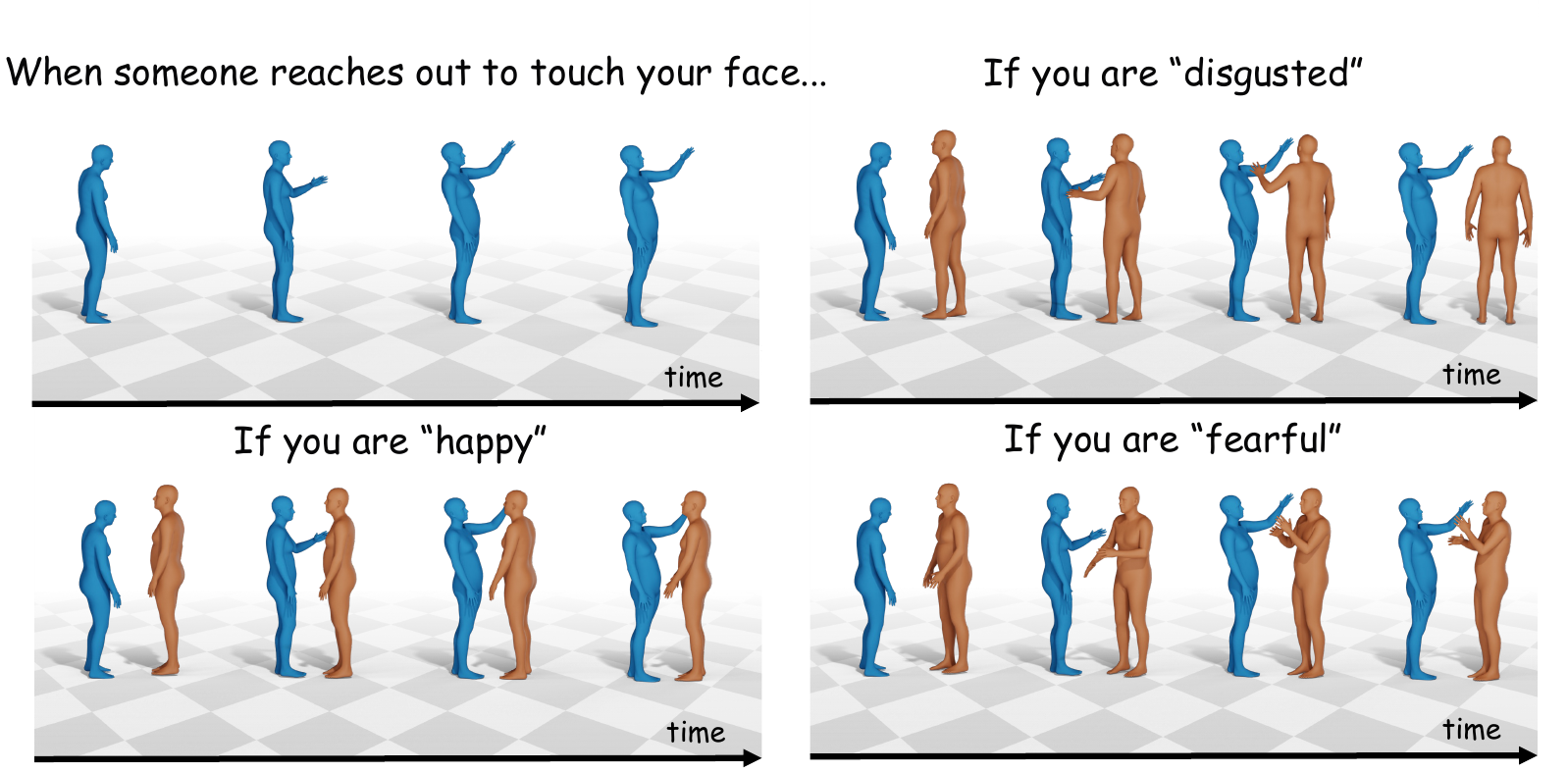}
  \vspace{-6mm}
  \caption{Our method generates diverse and realistic reactive motions conditioned on the actor’s movement and various reactor's emotion. The \textbf{Actor} and \textbf{Reactor} are represented in \textbf{blue} and \textbf{brown}, respectively.}
  \label{fig:teaser}
  \vspace{-5.5mm}
\end{figure}

To explicitly consider emotions, we introduce a novel task of emotion-aware reactive motion generation based on an actor’s movement and diverse emotional signals. As an emerging research area, it faces two major challenges. The first one is the lack of expressive emotion representation and data-efficient learning framework to achieve high-quality generation with limited emotion-conditioned motion data. Another critical obstacle is to maintain both emotional expressiveness and spatial coherence in two-person interaction scenarios. Existing reaction generation frameworks~\cite{xu2024regennet, ghosh2024remos} typically encode the actor's motion into implicit embeddings, which may fail to preserve the geometric structure and interaction constraints, often resulting in distorted interaction dynamics.

To address these challenges, we propose the first emotion-driven human reaction generation framework, named \textbf{E-React}. To reduce the dependency on labeled data, we introduce a semi-supervised learning mechanism to learn an expressive emotion prior for 7 emotion types. Based on the observation that motion clips within a short sequence tend to share the same emotion, the framework can capture high-dimensional affective semantics from a large amount of unlabeled data with a prediction task. To further enhance interaction quality, we design a symmetrical actor-reactor denoising architecture, which explicitly incorporates spatial relationship awareness into the diffusion process. By conditioning the diffusion model on the learned emotion prior, our framework facilitates both emotional expressiveness and spatial coherence in reactive motions.

\begin{figure*}
  \centering
  \includegraphics[width=1.0\textwidth]{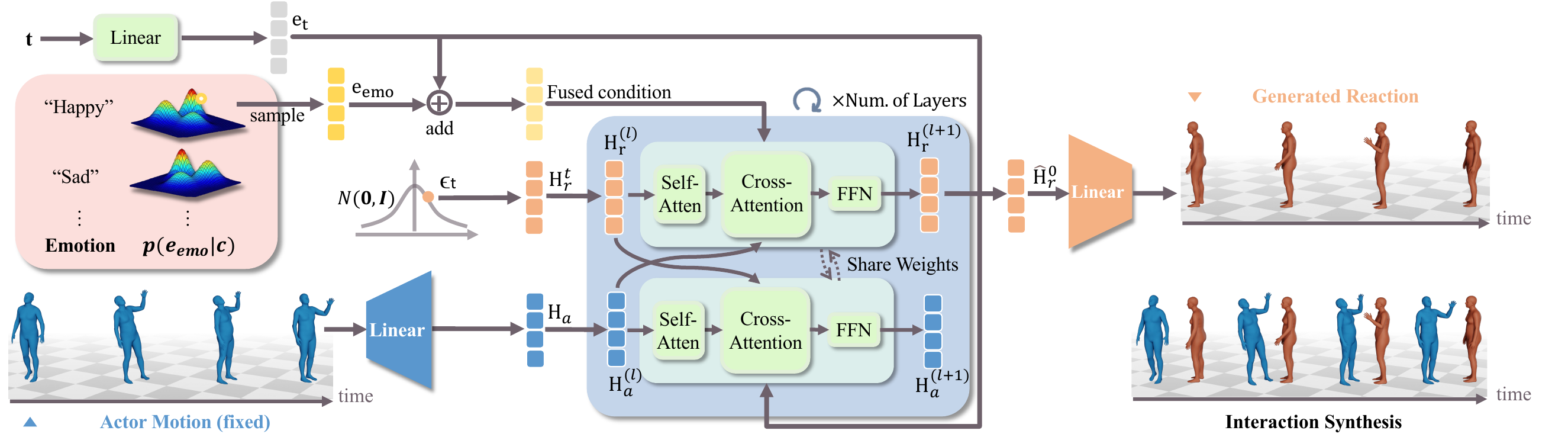}
  \vspace{-6mm}
  \caption{\textbf{Overview of the proposed E-React framework.} Given a specified emotion type, we first sample an emotion embedding from a semi-supervised emotion prior, which captures emotion-related dynamics in motion sequences. Subsequently, the actor's motion and a sampled noise are fed into a symmetrical diffusion model, where the actor's motion is fixed to maintain the spatial consistency of 3D motion. After several diffusion timesteps, the model progressively refines the noise into a temporally coherent reactive motion, guided by the emotion embedding, to generate a natural and expressive response.
  }
  \label{fig:pipeline}
  \vspace{-4mm}
\end{figure*}

Specifically, we first learn a probabilistic emotion prior by disentangling emotional features from 3D human motions through an emotion recognition task. Since no existing motion-emotion annotated dataset is available, we annotate 2500 motion sequences from the Inter-X dataset~\cite{xu2024inter} with emotion labels~(e.g.\ happiness and anger). By combining this annotated data with large-scale unlabeled motions, we train the network to classify motions into different emotion categories through a semi-supervised learning paradigm. After training, we extract the emotional prior from learnable emotional tokens using a clustering algorithm. This prior is modeled as multiple Gaussian distributions, with each distribution representing a distinct emotion. Compared to fixed label embeddings~\cite{degardin2022generative, petrovich2021action}, this probabilistic representation captures diverse and continuous emotional characteristics, enabling more expressive and flexible emotion-driven reaction generation. During the reaction generation, we first sample an emotion embedding from a specific Gaussian distribution, and then synthesize a reactive motion conditioned on the given actor motion and the sampled embedding. As illustrated in Figure~\ref{fig:pipeline}, unlike previous works~\cite{xu2024regennet, ghosh2024remos} that directly use implicit motion embeddings as conditions, we design a symmetrical actor-reactor architecture to explicitly preserve spatial information. 

The reactor's motion is reconstructed from a sampled noise vector, while the actor's motion remains clean and fixed throughout the entire diffusion process.  Consequently, the original geometric structure and interaction constraints in the actor motion can be transferred to the reactor through a cross-attention module, which facilitates consistent reaction generation. Since the reaction generation task requires the reactor motion to respond to the actor's motion, this symmetrical architecture is better suited than previous networks~\cite{shafir2024human, tanaka2023role, liang2024intergen}. Under the guidance of the sampled emotion embedding, emotionally expressive and natural reactive motions are progressively generated through several diffusion timesteps. To ensure that the generated motions align with the intended emotional signals, we further employ the emotion encoder to extract emotion embeddings from the generated motions. These embeddings are then supervised to match the input emotion embedding, enabling our framework to produce diverse and realistic reactions consistent with different emotional inputs. In summary, our contributions are as follows:

\begin{itemize}

\item We propose the first emotion-driven human reaction generation framework and reveal that affective signals can significantly improve the realism and diversity of human interaction generation.

\item We devise a emotion prior to learn implicitly affective knowledge from limited annotations with a semi-supervised mechanism.

\item We introduce a symmetrical actor-reactor denoising architecture to enhance spatial relationship awareness, which facilitates a natural interaction generation.

\end{itemize}

\section{Related Work}
\label{sec:relatedwork}

\noindent\textbf{Emotion Recognition.} Emotion recognition enables intelligent agents to communicate with humans in a natural and authentic manner. 
In this field, deep learning-based approaches are broadly used, which typically performed 7-class emotion classification after preprocessing stages such as dimensionality reduction and keyframe extraction~\cite{sapinski2019emotion}. The 7-class scheme originates from psychological research~\cite{ekman1992argument}, which identifies 6 basic emotions, with an additional "neutral" class commonly used in affective computing. Based on this design, an attention-based LSTM~\cite{zhang2021emotion} improved the recognition by identifying more significant body joints for affective expression and providing different position weight matrices. Recent work has also proposed graph representation~\cite{ghaleb2021skeleton} and multi-scale temporal-spatial analysis in affect recognition tasks~\cite{wang2023emotion, ouguz2024emotion}. However,the majority of research in emotion recognition has focused on the analysis of speech~\cite{sonmez2020speech, 10.1007/s00138-022-01294-x, 10097258, 10095388}, and facial expressions~\cite{Zou2021ANM, yan2022hybrid, singh2023facial}, which neglect the prior information in the skeletons and lead to inferior recognition performance. In contrast, we propose a semi-supervised framework and decompose the emotion information from human motions, which expands the application scope of emotion recognition.

\noindent\textbf{Conditional Motion Synthesis.} Conditional human motion generation leveraging technologies like GANs~\cite{habibie2022motion}, VAEs~\cite{guo2020action2motion}, and Diffusion models~\cite{tevet2023human, ao2023gesturediffuclip, liang2024omg} has gained significant attention. Control signals typically include action labels~\cite{degardin2022generative, guo2020action2motion}, textual descriptions~\cite{ghosh2021synthesis, zhang2022motiondiffuse, lee2024t2lm}, music~\cite{petrovich2021action, kim2022brand}, speech~\cite{zhu2023taming, chhatre2024emotional}, scene objects~\cite{wang2021scene, huang2023diffusion, diller2024cg}, and incomplete motion sequences~\cite{harvey2020robust, duan2021single, cohan2024flexible, wu2024finger}. However, human behavior is often driven by emotions, and only few works have focused on emotion-driven motion generation. For instance, \cite{liu2022beat} introduced a Cascaded Motion Network, which integrates facial expressions and emotions for the generation. Gesturediffuclip~\cite{ao2023gesturediffuclip} used large language models to generate emotional prompts for stylized gesture generation. Additionally, \cite{zhang2024large} proposed a large-scale model using emotional tags to aid in generating gestures and facial expressions from speech. However, most of these studies focus on speech-to-gesture tasks and overlook emotion expression through full-body movements. Moreover, bridging modalities with large models requires extensive data and computational resources~\cite{chen2025language, lin2023motion}. The fusion of control signals across different modalities remains an open challenge. Our model addresses these gaps by integrating emotional control and motion information from interacting individuals, which leads to more realistic human motion generation.

\noindent\textbf{Multi-Person Motion Synthesis.} Synthesizing the movements of multiple people in the same scene is a challenging yet valuable research task. Previous work has focused on multi-person motion prediction~\cite{adeli2020socially, guo2022multi, tanke2023social, jeong2024multi}, which often use interaction data to improve the accuracy of future movement predictions. In addition, the concept of Action-Reaction was first introduced in~\cite{huang2014action}, where they modeled the interaction as an optimal control problem, predicting one agent’s reaction based on the other's actions. However, the reactions to an actor's motion are not always deterministic. Recent works have used generative models to introduce variability in interactions~\cite{xu2023actformer, chopin2024bipartite, ghosh2024remos}. ReGenNet~\cite{xu2024regennet} proposes an online model that generates responses in SMPL-X~\cite{SMPL-X:2019} format based on input actions. Contact information~\cite{gu2024contactgen} is also used as constraints to improve accuracy. Additionally, some works~\cite{tanaka2023role, liang2024intergen} use textual descriptions to guide two-person motion generation, but these methods only provide semantic guidance for overall sequences and struggle to produce natural close-proximity interactions. In contrast, our approach introduces an actor-reactor diffusion model to combine the implicit features from both the actor and reactor during the denoising process, allowing for more precise interaction synthesis.

\vspace{-1.5mm}
\section{Method}

\vspace{-0.2mm}
Given a motion of an actor, our goal is to drive a reactor to generate diverse, authentic, and natural reactions with different emotions.
To this end, we first train an emotion prior to decompose implicit emotional features from human motions with limited annotated data. With the trained prior, we further propose an actor-reactor diffusion model guided by emotion embedding to learn spatially interactive motions through a symmetrical structure. Finally, given a motion sequence of an actor, our approach can generate realistic reactions under various emotional conditions.

\subsection{Problem Formulation}
\label{subsec: Problem Formulation}
Following previous action-conditioned generation methods~\cite{ghosh2024remos,xu2024regennet}, we define the character with known movements as the \textbf{actor}, and the character whose motion we aim to generate as the \textbf{reactor}.
The motions of the actor and reactor are represented as $X_a=\{x_a^i\}_{i=1}^L$ and $X_r=\{x_r^i\}_{r=1}^L$, respectively, where $L$ is the frame length.

To represent interactions, the two characters are modeled in a unified coordinate system, following GMD~\cite{karunratanakul2023guided}. For each frame, we adopt the non-canonical motion representations to facilitate network learning. Specifically, person $p$ in frame $i$ is represented as:
\begin{equation}
x_p^i=\left[j_p^i, v_p^i, r_p^i, f_p^i\right],
\end{equation}
where $j_p^i \in \mathbb{R}^{3N}$, $v_p^i \in \mathbb{R}^{3N}$, $r_p^i \in \mathbb{R}^{6\times(N-1)}$, $f_p^i \in \mathbb{R}^{4}$ are global joint positions, global joint linear velocities, 6D representation of local rotations in the root frame, and foot contacts, respectively. $N$ is the number of body joints.

With this representation, our task is to train a network $F$ to predict a reactive motion based on the actor's motion $X_a$, conditioned on a given emotional signal $e_{emo}$:
\begin{equation}
 X_r = F \big(X_a,e_{emo}\big).
\end{equation}

\begin{figure*}[t]
  \centering
    \centering
    \includegraphics[width=1.0\textwidth]{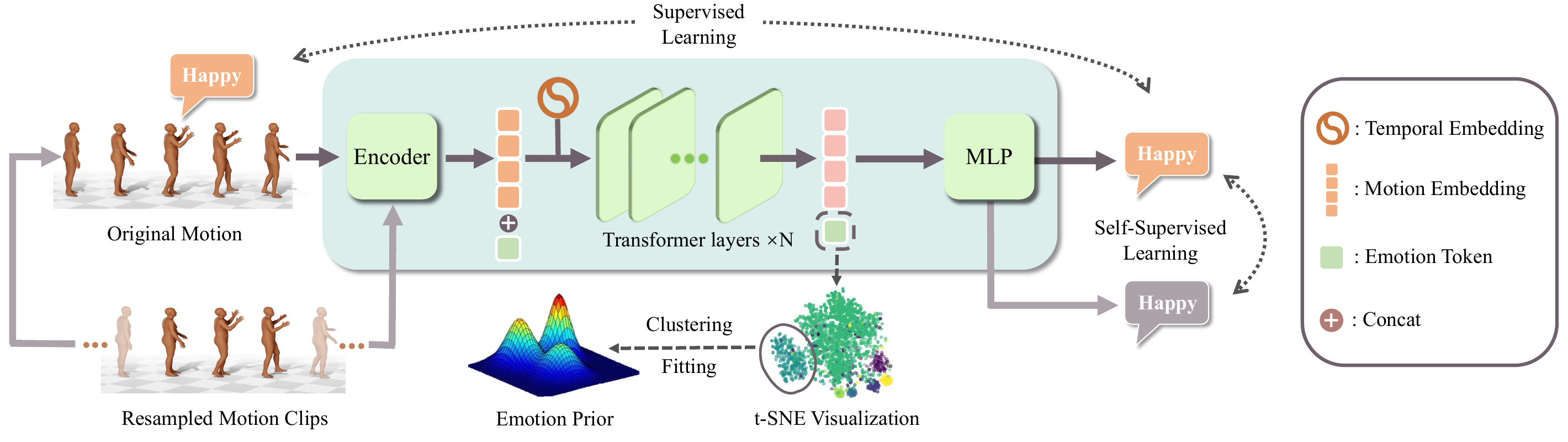}
    \vspace{-6mm}
    \caption{Illustration of the semi-supervised training process. For motion sequences with emotional annotations, we employ ground-truth label to supervise the predicted results. For unlabeled motion data, we first generate multiple shorter motion clips through repeated sampling, and then train the model using emotion consistency constraints across these clips.}
    \label{fig:aprd_pipline}
    \vspace{-4mm}
\end{figure*}

\subsection{Semi-supervised Emotion Prior}
\label{subsec: Emotion Feature Encoder}
Since emotion is abstract and ambiguous, it cannot be explicitly integrated into reaction generation framework. To this end, we propose a semi-supervised network to learn an emotion prior in the latent space.

\noindent\textbf{Model architecture.} We first design a network to classify motions into different emotion categories. The motion sequence $X$ is projected into a latent vector $E_m =\{e^i\}_{i=1}^L \in \mathbb{R}^{L*D_e}$ via a linear layer. A learnable classification token $e_{cls} \in \mathbb {R}^{D_e}$ and a learnable temporal embedding $E_T \in \mathbb {R}^{(L+1) \times D_e}$ are also concatenated to capture effective features. Consequently, the motion features is represented as:
\begin{equation}
    E = \left[e_{cls}, E_m\right] + E_T , \in \mathbb {R}^{(L+1) \times D_e},
\end{equation}
where $D_e$ is the dimension of the embedded feature space, and $Linear$ represents the learnable linear projection layer. Subsequently, a cascaded transformer layer based on a multi-head self-attention mechanism is used to extract information at different scales. The operation in layer $l$ can be expressed as:
\begin{equation}
\begin{aligned}
    E^{(l+1)} & =E^{(l)} + Self(Ln(E^{(l)}))  \\
              &+ Mlp( Ln( E^{(l)}+ Self(Ln(E^{(l)})) ) ),
\end{aligned}
\end{equation}
where $Ln$ denotes layer normalization, and $Mlp$ represents the multi-layer perceptron block. $Self$ refers to the multi-head self-attention layer. 

After being updated through $N_l$ layers of Transformer, the emotion token can learn information from the entire motion sequence. We thus take the updated token $e_{emo} = e^{(N_l)}_{cls}$ as the emotion embedding, which is then fed into a $Mlp$ block to predict the emotional category:
\begin{equation}
    P_{emo} = Mlp(e_{emo}), \in \mathbb {R}^{D_{cls}},
\end{equation}
where $P_{emo}=\{ p_c \}_{c=1}^{D_{cls}}$ is a one-hot vector representing the predicted emotional probability, and $D_{cls}$ denotes the number of emotional label categories.

\noindent\textbf{Semi-supervised training.} Since no existing emotion-motion paired data can be used to train the network, we first annotate a small set of 2500 motion sequences in Inter-X dataset~\cite{xu2024inter}. However, it is still insufficient to learn an expressive emotion prior. Fortunately, we find that motion clips within a short sequence tend to share the same emotion. We thus propose a semi-supervised manner to leverage large-scale motion dataset and the partially labeled annotations to train our prior. As shown in Figure~\ref{fig:aprd_pipline}, we resample $k$ motion clips from the same sequence, and then fed them into the network. Since these clips have the same emotional states, we enforce their emotion tokens to be consistent:

\begin{equation}
    L_{con} = \|e^i_{emo}-e^j_{emo}\|_2^2,
\end{equation}
where $e^i_{emo}$ and $e^j_{emo}$ are emotion tokens of the clips sampled from the same sequence.

For the sequences that have emotion labels, we directly supervise the network output with a Cross Entropy Loss:

\begin{equation}
\begin{aligned}
    L_{ce}= & \frac{1}{N}\sum_{i}L_{i}=-\frac{1}{N}\sum_{i}\sum_{c=1}^{D_{cls}}y_{ic}\log(p_{ic}), \\
y_{ic}= &
\begin{cases}
1,\ {emo}_i\in c \\
0,\ {emo}_i\notin c 
\end{cases}
\end{aligned}
\end{equation}
where $emo_i$ denotes the emotional label of motion clip and $p_{ic}$ represents the predicted probability that $emo_i$ belongs to category $c$. 
The overall training loss for the emotion prior is:
\begin{equation}
    L_{prior}=  L_{ce} +  L_{con}.
\end{equation}

$L_{ce}$ enables our model to learn a sparse emotional feature distribution from limited annotated data. Through resampling extensive emotion-unlabeled data and applying consistency loss $L_{con}$ supervision, we effectively reduce the distance between emotionally consistent samples in the high-dimensional space.  

This training strategy enables the model to learn a more expressive emotion representation from large-scale unannotated motion data.

Once the training is completed, we encode all motion sequences to obtain emotion tokens. With these tokens, we adopt a clustering algorithm and Gaussian distributions to model the prior for each emotion category. We first calculate the initial mean and variance of the tokens from the annotated data for 7 emotion categories. These computed means serve as class centers for clustering the tokens from all motion sequences. Subsequently, we fit a Gaussian distribution to each obtained cluster, ultimately producing 7 distinct Gaussian distributions. Based on the clustering of dense emotional features, these distributions constitute an effective emotion prior, from which we can sample emotion embeddings for each emotion category.

\subsection{Symmetrical Actor-Reactor Motion Denoising}\label{subsec: Independent Actor-Reactor Motion Denoising}
With the learned prior, we can train an actor-reactor diffusion model for reaction generation as shown in Figure~\ref{fig:pipeline}. Our objective is to establish a conditional probability distribution $p(X_r|X_a, \hat{e}_{emo})$ from which natural reactive movements can be sampled, where $\hat{e}_{emo}$ is an embedding sampled from the learned prior. 

In this section, we present our actor-reactor motion diffusion model that employs a divide-and-conquer strategy to separately process low-level condition~(spatial relationship) and high-level condition~(emotion). This framework simultaneously enables natural interaction generation and effective emotional control.

\noindent\textbf{Human Emotional Reaction diffusion.} Diffusion models have been proven to possess powerful generative capabilities, enabling them to construct corresponding distributions from complex data features. We have constructed a reaction generation framework based on diffusion models, as illustrated in Figure~\ref{fig:pipeline}. Following~\cite{ho2020denoising}, during the forward process, we gradually add $T$ steps of Gaussian noise $\epsilon\sim\mathcal{N}(\mathbf{0},\mathbf{I}) $ to ground-truth reaction motion $X_r$, resulting in $X^{(T)}$ that approximately follows a standard normal distribution:
\begin{equation}\label{equation:forward_diffusion_origin}
    X^{(t)} = \sqrt{\hat{\alpha}_t}X^{(0)} + \sqrt{1 - \hat{\alpha}_t} \epsilon, \epsilon \sim \mathcal{N} \left(0, \rm{I} \right),
\end{equation}
where $t$ denotes the timestep, and $\bar{\alpha}_t=\prod_{i=1}^t\alpha_i$, $a_i \in(0,1)$ is a constant hyper-parameter. In the denoising process, we follow previous work~\cite{tevet2023human, karunratanakul2023guided, xu2024regennet} to employ a network $F$ to directly estimate the clean reactive motion sequence conditioned on actor motion and emotion embedding. Compared to estimating noises, this strategy makes the framework easier to apply spatial constraints.
\begin{equation}
    X_r^{(0)}= F \big(X_r^{(t)},t,X_a,\hat{e}_{emo}\big).
\end{equation}

\noindent\textbf{Symmetrical Actor Motion Branch.} Previous works~\cite{ghosh2024remos, xu2024regennet} simply employ an independent encoder to obtain embeddings of the actor's motion, utilizing them as external conditions to the diffusion process. This approach results in weak spatial relationships, especially in scenarios with multiple modality conditions~(e.g.\ motion and emotion), leading to multi-modal conditional aliasing. 
Since both action and reaction are motion sequences, we propose a symmetric architecture that directly establishes their perceptual relationship within the diffusion framework.
The actor's motion is projected into the same latent space using shared-weight network of the reactor motion, thereby facilitating the perception of robust spatial relationships.

Notably, we maintain the input actor's motion as fixed and clean throughout the process, while directly connecting it to the reactor's denoising process via cross-attention mechanisms:
\begin{equation}
\begin{aligned}
    H^{(l+1)}_r &= Atten\big(H^{(l)}_r, H^{(l)}_a, t, \hat{e}_{emo}\big) \\
    H^{(l+1)}_a &= Atten\big(H^{(l)}_a, H^{(l)}_r, t\big),
\end{aligned}
\end{equation}
where $H^{(l)}_a, H^{(l)}_r\in \mathbb {R}^{D_{L}}$ and represent the intermediate motion embeddings of the actor and reactor respectively, after $l$ layers of transformer decoder.

During the forward and reverse diffusion processes, the feature of action is always the clean embedding without any noise on it. With robust spatial perception capability of our actor-reactor architecture, we generate more natural and realistic reaction while maintaining precise emotional control. Ablation study have demonstrated that such design yields enhanced results.

\noindent\textbf{Learning objective.} We utilize a reactive loss function $L_{react}$ composed of the joint distances between two individuals to supervise the diffusion model.
\begin{equation} 
    L_{react} = \|J(\hat{\mathbf{X}}_{a},\mathbf{X}_{r}) - J(\hat{\mathbf{X}}_{a},\hat{\mathbf{X}}_{r})\|_{2.}^{2.},
\end{equation}
where $J(X_a, X_r)$ presents the joint distances between actor and reactor, and $\hat{X}$ denotes the ground truth motions.

To ensure the generated reaction sequences are reasonable motions, we utilize a common geometric loss $L_{geo}$ in human regularization, which encompasses bone lengths, velocity smoothness, and foot contact consistency. These loss functions are the same as previous works~\cite{liang2024intergen, tevet2023human, xu2024regennet}, and the details can be found in the sup. mat. 

We further introduce an emotional loss function $L_{emo}$ to enforce the emotion embeddings of generated reactions to align with the corresponding distribution of the input emotion type:
\begin{equation} 
    L_{emo} = \frac{(\hat{e}_{emo}-\mu)^2}{2\sigma^2},
\end{equation}
where, $\mu$ and $\sigma$ are the mean and variance of a specific Gaussian distribution. This loss can promote the network to generate motions with diverse emotions. Overall, our learning objective can be written as:
\begin{equation}
    L_{total}  =  L_{rc} +  L_{react} + L_{bone} 
     +   L_{smooth} +  L_{foot} + L_{emo}.
\end{equation}

\subsection{Emotion-driven Reaction Generation}\label{subsec: Emotion-driven Human Reaction Diffusion}

With the emotion prior and symmetrical architecture, we can achieve the following applications.

\noindent\textbf{Empathetic Reaction Generation.}
In social psychology, human interactions in the real world are often accompanied by a phenomenon known as emotional contagion. 
Given an actor motion with an unknown emotional state, we can employ the encoder of our emotion prior to predict the actor's emotion $\hat{c}$, and then sample $\hat{e}_{emo}$ from the distribution corresponding to $\hat{c}$. Consequently, we can control the framework to generation reactions that are emotionally consistent with the actor's inferred state.
We perform a comparative analysis between empathetic reaction generation and unconditional reaction generation, with detailed results and analysis provided in the sup. mat.

\begin{figure*}
  \centering
  \includegraphics[width=1.0\textwidth]{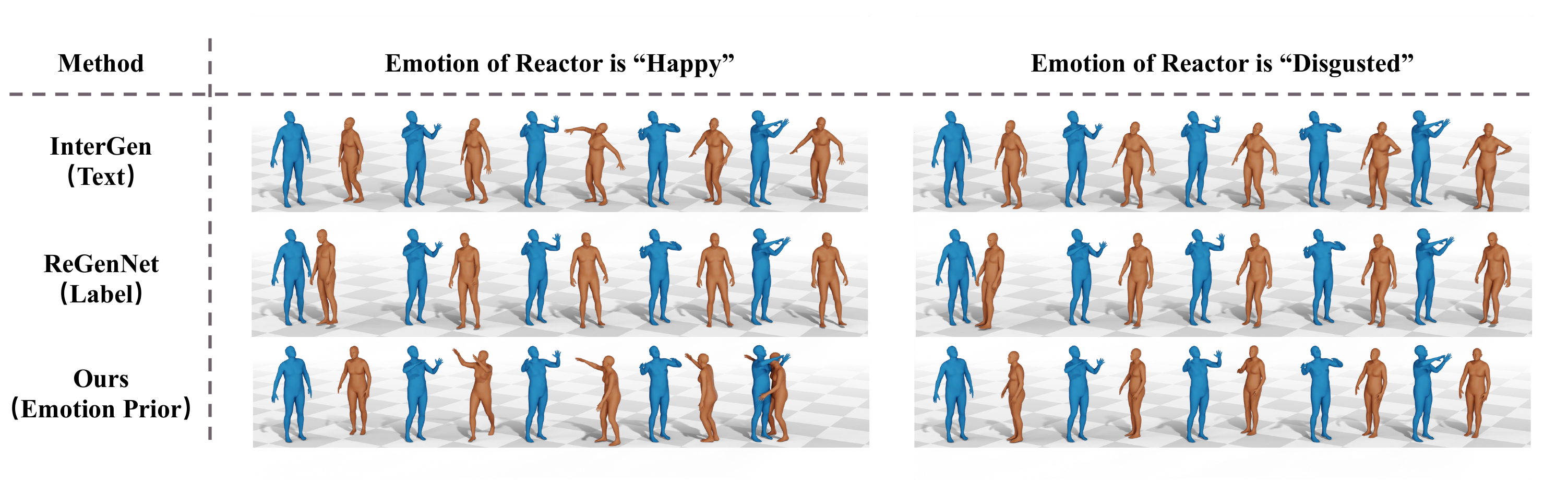}
  \vspace{-8mm}
  \caption{Qualitative comparison of emotion-guided human reaction synthesis. The actor (blue) is making joyful gestures towards the reactor (brown). Base on InterGen, ReGenNet and our own model, we synthesize human reactions under the emotional conditions of "happy" and "disgusted". }
  \label{fig:comparsion}
  \vspace{-5mm}
\end{figure*}

\noindent\textbf{Emotional Editing of Reactions.} We can also generate different reactive motions with diverse emotion states to response to a specific actor motion. During the inference process, we sample an implicit emotion embedding $\hat{e}_{emo}$ from our emotion prior $p_c(e_{emo})$ according to a specific emotion type. With the embedding and actor motion, our framework can produce realistic emotional reactions as shown in the last row of Figure~\ref{fig:comparsion}. More quantitative results on emotion-driven performance can be found in sup. mat.

\section{Experiments}
\label{Experiments}

\noindent\textbf{Dataset.} Due to the limited emotion labels, we manually annotated the emotions for multi-person motion datasets, Inter-X~\cite{xu2024inter}. We named the annotated dataset as \textbf{Inter-X-e}. Inter-X-e contains 11,388 motion sequences and 2,500 emotional tags. Among them, 500 sequences with emotion labels are used for evaluation, while the rest are used for training. Our semi-supervised learning framework utilizes 2,000 labeled and 7,000 unlabeled motions to train the emotion prior. With the trained prior, we train the actor-reactor diffusion model on the standard training split of Inter-X. More details of the emotion annotation process can be found in sup. mat.
 
\noindent\textbf{Metrics.} Follow recent advancements in motion generation works~\cite{liang2024intergen, ghosh2024remos}, we employ FID~\cite{heusel2018ganstrainedtimescaleupdate} to measure the distribution gap in latent space between the generated and ground-truth reaction. We also implement Diversity~(DIV) to assess the latent diversity of generated reactions, and Multimodality~(MM) to evaluate the latent diversity within the same emotional category. We further introduce Accuracy~(ACC) $ACC=T/N$ to evaluate the consistency between the emotion of generated reactions and given emotional conditions, where $T$ represents the number of generated samples whose predicted emotion~(obtained via a frozen emotion prediction network) matches the input emotional condition, and $N$ is the total number of generated samples. 

\subsection{Comparison with State-of-the-art}
\label{subsec: Comparison}

Since previous works on reaction generation do not consider emotional inputs, we have made modifications on some conditional motion generation frameworks~\cite{xu2024regennet,liang2024intergen}. Specifically, for ReGenNet~\cite{xu2024regennet}, a reaction generation framework conditioned on action labels, we adopt SMPL representation and replace the action label with emotion label encoded as one-hot vectors. Consequently, this framework can also generate emotion-conditioned reactions. In addition, InterGen~\cite{liang2024intergen} generates two-person interactions according to specific texts. We use textual descriptions such as "happy", "disgusted" and "sad" encoded with CLIP~\cite{radford2021learning} as a condition, which can be considered as a high-level emotion signal.

\noindent\textbf{Qualitative Comparisons.} The qualitative comparisons with these two baseline methods are presented in Figure~\ref{fig:comparsion}, where the actor and the reactor are represented in blue and brown, respectively. As shown in the first row, the actor makes joyful gesticulations towards the reactor. Then we demonstrate two emotional states: happy and disgusted.
In ReGenNet, we regard emotion as a distinct action category, representing it through a one-hot vector that is subsequently encoded into an embedding space via a linear layer. The emotion embedding is then concatenated with the actor's motion to form the condition input. The ReGenNet framework fails to generate plausible reaction motions due to its limited capacity to perceive spatial relationships and capture emotional information. Although InterGen understands emotional textual description with CLIP's knowledge, it still demonstrates limited capability in capturing interactive intentions. In addition, InterGen simultaneously predicts two-person motions, which cannot produce natural reactions. Furthermore, constrained by the limited expressiveness of emotional textual descriptions, the framework exhibits insufficient diversity across different emotional conditions.

In contrast, in our method, the reactor generated by our model waves its arms enthusiastically and attempts to step forward to embrace the actor when the reactor's emotion is specified as "happy". When the reactor's emotion is set as "disgusted", the output of our model simply gives a slight wave of the hand and performs a body-turning motion. Our results show obvious interactive behavior. Besides, the "disgusted" reaction significantly contrasts with the "happy" condition's attempt to shorten social distance, indicating that our model can generate corresponding actions based on different emotions and produce diverse outcomes.

\noindent\textbf{Quantitative evaluation.} The quantitative results in Table~\ref{tab: conditional quantitative comparison} demonstrate the effectiveness of our model in generating emotional reaction. Due to the spatial information of the actor extracted by the symmetrical design, our method can achieve better performance in terms of FID. Meanwhile, our network effectively preserves the diversity of reaction generation. Although InterGen is comparable to our model in MM metric, its performance on the FID is noticeably poorer, indicating that it is inferior in reaction generation task. With the emotion representation in emotion priors, the ACC reflects that our model can effectively control the emotion in reactions. Furthermore, we also compared to existing reaction generation methods without additional emotional control to demonstrate the superiority of our framework in modeling spatial interaction relationships. We retrain both our model and all baseline models conditioned on the actor's motion. The results of the comparison are shown in the sup. mat.

\begin{table}
\centering
\resizebox{\linewidth}{!}{
  \begin{tabular}{@{}ccccc@{}}
    \toprule
    Method & FID $\downarrow$ & DIV $\rightarrow$ & MM $\uparrow$ & ACC $\uparrow$\\ \midrule
    GT &$0.79^{\pm{0.03}}$ &$6.06^{\pm{0.02}}$ &$-$ &$0.89$ \\
    InterGen &$5.54^{\pm{0.07}}$ &$4.54^{\pm{0.05}}$ &$1.63^{\pm{0.02}}$ &$0.67$ \\
    ReGenNet &$3.82^{\pm{0.03}}$ &$4.28^{\pm{0.02}}$ &$1.34^{\pm{0.04}}$ &$0.59$ \\
    \textbf{Ours} & $\mathbf{1.94}^{\mathbf{\pm{0.03}}}$ & $\mathbf{5.74}^{\mathbf{\pm{0.06}}}$ & $\mathbf{1.71}^{\mathbf{\pm{0.02}}}$ &$\mathbf{0.86}$ \\ 
    \bottomrule
  \end{tabular}
  }
  \vspace{-1mm}
  \caption{Quantitative comparison on Inter-X-e dataset for emotion-conditioned human reaction generation task.}
    \label{tab: conditional quantitative comparison}
\end{table}

\subsection{Ablation Study}
\label{subsec: Ablation}

\noindent\textbf{Semi-supervised mechanism.}
To validate the effectiveness of semi-supervised mechanism, we compared the model performance of the following training configurations: (1) supervised training with all the 2000 labeled samples, and (2) semi-supervised training with 2000 labeled samples and additional $N$ unlabeled samples from the original 11,388 actions, where $N = 2000, 5000, 7000, 9000, all$. 

In Table~\ref{tab:ablation of semi-supervise}, the results demonstrate that our semi-supervised mechanism significantly enhances emotion prediction accuracy by effectively leveraging unlabeled data, thereby facilitating more effective emotional modeling. The prediction accuracy improves significantly as the number of self-supervised samples increases. However, excessive unlabeled data causes model predictions to be biased toward the mean, leading to a decline in performance. When the prediction network approaches its performance plateau at $ N = 7000 $, we freeze the model weights at this configuration. The frozen model is subsequently employed for both constructing our emotion prior and computing the emotional accuracy of generated reactions.

\begin{table}
\centering
\resizebox{\linewidth}{!}{
  \begin{tabular}{@{}l|cccccc@{}}
    \toprule
    Unlabeled samples $N$ & Supervised & 3000 & 5000 & \textbf{7000} & 9000 & all \\ \midrule
    Accuracy $\uparrow$ & 0.71 & 0.81 & 0.86 & \textbf{0.89} & 0.83 & 0.84 \\
    \bottomrule
  \end{tabular}
  }
  \vspace{-1mm}
  \caption{Ablation on the necessity of semi-supervised mechanism and the effect of unlabeled sample size $N$.}
  \label{tab:ablation of semi-supervise}
    \vspace{-5mm}
\end{table}

\noindent\textbf{Emotion Prior.} To validate the emotion prior, we performed a stepwise ablation study. We use 3 different emotion condition as input. The first is a one-hot encoded vector representing an emotion category. The second is the centroid of clusters formed by learned emotion embeddings. The third is an emotion embedding sampled from our trained prior. In Table~\ref{tab:ablation on the effect of emotion prior}, we found that only the categorical labels represented by one-hot vectors are insufficient for both accurate emotion modeling and the generation of emotion-specific reactions~(row 2).
The results demonstrate that our prediction network captures discriminative emotional features. The emotion prior exhibits superior expressive capability for emotional representation~(row 3). With a sampling strategy, our method can further improve diversity and accuracy~(row 4). These findings collectively confirm the necessity of the emotion prior for our reaction generation task. 

\begin{table}
\centering
\resizebox{\linewidth}{!}{
  \begin{tabular}{@{}ccccc@{}}
    \toprule
    Condition & FID $\downarrow$ & DIV $\rightarrow$ & MM $\uparrow$ & ACC $\uparrow$ \\ \midrule
    GT &$0.79^{\pm{0.03}}$ &$6.06^{\pm{0.02}}$ &$-$ &$0.89$  \\
    One-hot &$8.91^{\pm{0.05}}$ &$1.46^{\pm{0.02}}$ &$0.95^{\pm{0.01}}$ &$0.52$ \\
    Cluster centroid &$2.34^{\pm{0.02}}$ &$5.17^{\pm{0.01}}$ &$1.58^{\pm{0.04}}$ &$0.81$ \\
    \textbf{Sampled embedding} & $\mathbf{1.94}^{\mathbf{\pm{0.03}}}$ & $\mathbf{5.74}^{\mathbf{\pm{0.06}}}$ & $\mathbf{1.71}^{\mathbf{\pm{0.02}}}$ &$\mathbf{0.86}$  \\ 
    \bottomrule
  \end{tabular}
  }
  \vspace{-2mm}
  \caption{Ablation on the effect of emotion prior module.}
  \label{tab:ablation on the effect of emotion prior}
  \vspace{-1mm}
\end{table}

\noindent\textbf{Symmetrical Actor-Reactor Denoising.}
Since the actor and reactor are expected to maintain a reasonable spatial relationship, our model employs an actor-fixed symmetrical denoising structure to preserve the actor's spatial information. We conducted experiments using an \textbf{asymmetric} architecture, where the actor's motion is processed through an independent self-attention network rather than a weight-sharing network with the reactor. We performed experiments with \textbf{non-fixed} actors, similar to generating dual-person motion from text. In these experiments, we applied noise to both the actor's and reactor's motions, and at each step of the denoising process, we replaced the predicted actor motion with ground truth values. 

We evaluate the interaction effectiveness of the generated results as shown in Table~\ref{tab: ablation on the effect of independent actor-reactor denoising structure.}. The actor-reactor denosing structure outperforms in FID, DIV and MM, indicating that the our design helps reinforce the awareness of interaction between humans and generates more reasonable interactive responses. The model's performance in ACC metric also demonstrates that this architecture effectively avoids the entanglement of control signals, successfully achieving emotion control while generating natural reactions.

\begin{table}
\centering
\resizebox{\linewidth}{!}{
  \begin{tabular}{@{}ccccc@{}}
    \toprule
    Framework & FID $\downarrow$ & DIV $\rightarrow$ & MM $\uparrow$ & ACC $\uparrow$\\ \midrule
    GT &$0.79^{\pm{0.03}}$ &$6.06^{\pm{0.02}}$ &$-$ &$0.89$   \\
    asymmetric &$5.13^{\pm{0.05}}$ &$4.34^{\pm{0.03}}$ &$1.13^{\pm{0.01}}$
    &$0.56$  \\
    non-fixed &$3.61^{\pm{0.01}}$ &$4.75^{\pm{0.01}}$ &$1.43^{\pm{0.02}}$ & 0.74\\
    \textbf{actor-reactor} & $\mathbf{1.94}^{\mathbf{\pm{0.03}}}$ & $\mathbf{5.74}^{\mathbf{\pm{0.06}}}$ & $\mathbf{1.71}^{\mathbf{\pm{0.02}}}$ &$\mathbf{0.86}$  \\ 
    \bottomrule
  \end{tabular}
  }
  \vspace{-1mm}
  \caption{Ablation on the effect of symmetric actor-reactor denoising structure.}
  \label{tab: ablation on the effect of independent actor-reactor denoising structure.}
  \vspace{-4mm}
\end{table}

\section{Conclusion and Discussion}

In this work, we introduce a novel and challenging task: emotion-conditioned reaction generation. Through our semi-supervised learning framework, we effectively address the challenge of limited labeled data and establish a robust emotional representation. By leveraging the  actor-reactor denoising architecture, we are able to synthesize natural and realistic interactions. Experimental results demonstrate that our approach outperforms existing methods in emotional reaction generation tasks. While E-react excels in generating emotional reactions, there remains room for further improvement. Data scale limitations and kinematic modeling constraints affect the it's diversity and physical realism. Future work will focus on enhancing emotional variety through expanded datasets with richer affective categories and improving physical realism by integrating physics-aware synthesis to address kinematic artifacts like foot sliding.

\bibliography{aaai2026}


\clearpage

\twocolumn[%
\begin{center}
    \LARGE \textbf{Supplementary Materials}
    \vspace{1em}
\end{center}
]
\setcounter{secnumdepth}{20}
\appendix

In the supplementary material, we first outline the protocol used for data annotation. Next, we provide in-depth discussions and additional details to demonstrate the effectiveness and efficiency of our method. Finally, we present several key user studies and show some visual results to highlight the strengths of our approach.

\section{Protocol of Emotion Annotation}
\label{supsec: Details on dataset extension}

In this section, we first introduce the definition of emotions and then describe the detailed methodology employed to extend the original Inter-X dataset. Psychological research~\cite{ekman1992argument} defines human emotion to have 6 basic types, a principle which is adopted by most emotion recognition datasets~\cite{lee2019context, dhall2011static}. Thus, our classification includes 7 categories with an additional label, "neutral". Consistent with prior works on emotion recognition~\cite{sapinski2019emotion}, we classify each individual in the dataset as expressing one of the following seven emotions: \textbf{anger, disgust, fear, happiness, neutral, sadness, or surprise}.

\begin{figure*}
  \centering
  \includegraphics[width=0.9\textwidth]{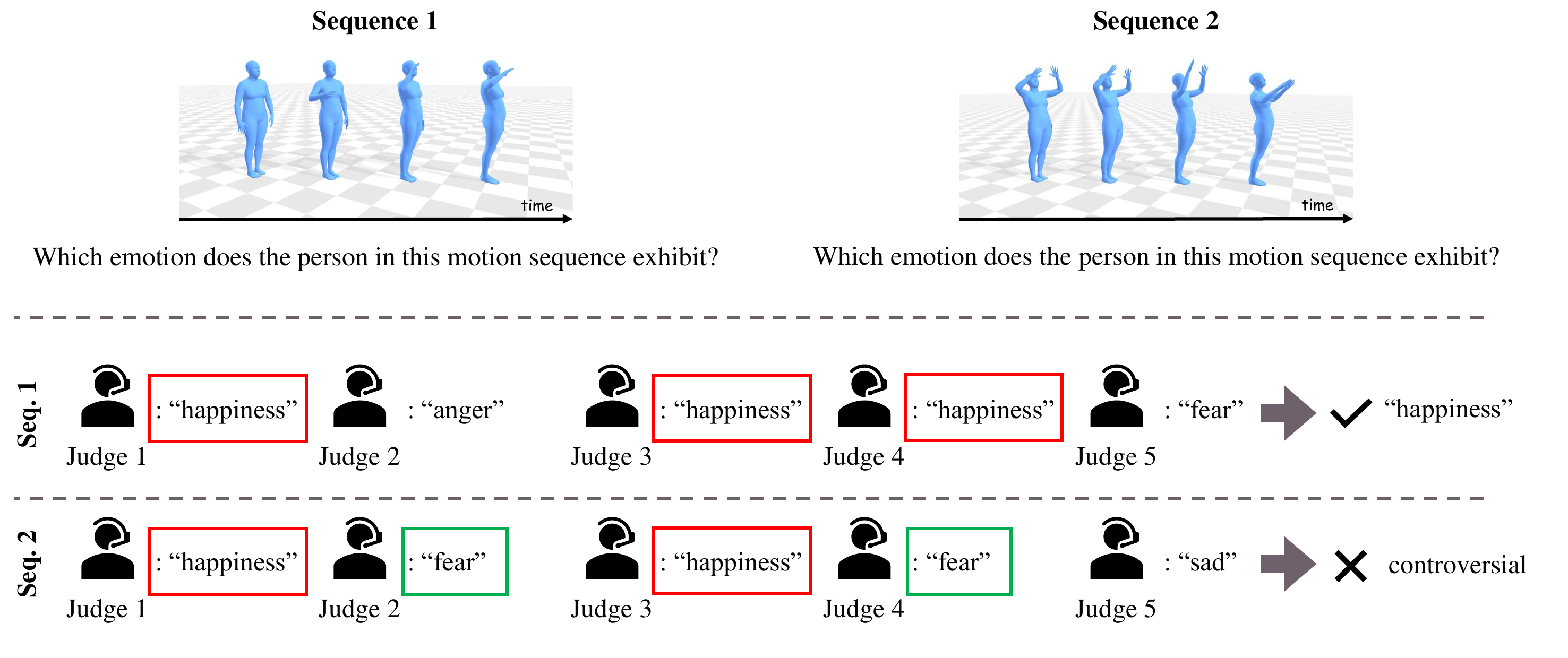}
  \caption{Illustration of the emotion annotation process.}
  \label{fig:sup_dataset}
\end{figure*}

Since identifying the emotion of a motion is challenging, we first develop a visualization tool that displays 3D motion sequences from different viewpoints. Subsequently, five annotators are tasked with reviewing the sequences and identifying the emotion type $c$ for each motion. The annotation process is illustrated in Figure~\ref{fig:sup_dataset}. If three or more annotators agree on the same category for a motion, the corresponding label is assigned to the sequence. As a result, we obtain 2,500 motion-emotion pairs, as summarized in Table~\ref{tab:dataset_num}.

To evaluate the annotations, we utilize the semi-supervised emotion prior network to classify the motions in a fully-supervised manner. The annotated dataset is split into 2,000 samples for the training set and 500 samples for the testing set. The results in Table~4 of the
main paper show that the network can effectively distinguish emotions. Additionally, we present the t-SNE visualization in Figure~5 of the main paper, where sequences with different emotional states exhibit similar characteristics in the motion space while displaying distinct emotion features in the latent space. These results demonstrate that emotions can be decomposed from body motions.

\section{More Discussion on Quantitative Results}
\label{supsec: Quantitative evaluation.}

Since no existing methods can generate reactions based on different emotional inputs, we perform a quantitative comparison with previous works by removing the emotional condition. Thus, the task becomes pure reaction generation. For different baseline methods, we replace the text condition in MDM~\cite{tevet2023human} with the actor's motion sequence. In InterGen~\cite{liang2024intergen}, we follow person-to-person setting and remove the text input. For ReMoS~\cite{ghosh2024remos}, we extract the only body joints for comparisons without extra setting~(e.g.\ hand-interaction awareness).

\begin{table}
\centering
\resizebox{\linewidth}{!}{
  \begin{tabular}{@{}l|ccccccc@{}}
    \toprule
    Type & neutral & happiness & anger & disgust & sadness  & fear  & surprise\\ \midrule
    Num &1680 &750 &490 &610 &575 &450 &445\\ \bottomrule
  \end{tabular}
  }
  \caption{The number of samples for different emotions in our Inter-X-e dataset.}
  \label{tab:dataset_num}
  
\end{table}

In contrast to the metrics for emotional reaction generation, when calculating the MM metric, we assess the diversity of generated reactions under the same actor motion.
As shown in Table~\ref{tab:unconditional Quantitative comparison}, our method leverages spatial information from the actor through the proposed actor-reator design, thereby achieving better performance in terms of FID. Furthermore, our network effectively preserves the diversity of reaction generation. Although InterGen surpasses our model in MM, one possible reason is that it inevitably introduces variations in the actor's motion during the denoising process, leading to higher diversity in reactions for the same action. Our method achieves state-of-the-art performance in both FID and DIV, demonstrating its ability to generate natural and diverse motions.

\begin{table}
\centering
\resizebox{\linewidth}{!}{
  \begin{tabular}{@{}lccc@{}}
    \toprule
    Method & FID $\downarrow$ & DIV $\rightarrow$ & MM $\uparrow$ \\ \midrule
    GT &$0.79^{\pm{0.03}}$ &$6.06^{\pm{0.02}}$ &$-$ \\ 
    MDM~\cite{tevet2023human} &$5.31^{\pm{0.03}}$ &$5.53^{\pm{0.06}}$ &$1.56^{\pm{0.02}}$ \\
    InterGen~\cite{liang2024intergen} &$3.32^{\pm{0.01}}$ &$5.35^{\pm{0.03}}$ &$\mathbf{1.67}^{\mathbf{\pm{0.04}}}$ \\
    Remos~\cite{ghosh2024remos} &$2.29^{\pm{0.04}}$ &$5.45^{\pm{0.01}}$ &$1.54^{\pm{0.05}}$ \\
    ReGenNet~\cite{xu2024regennet} &$2.13^{\pm{0.03}}$ &$5.62^{\pm{0.03}}$ &$1.53^{\pm{0.04}}$ \\ 
    \textbf{Ours} & $\mathbf{1.95}^{\mathbf{\pm{0.03}}}$ & $\mathbf{5.72}^{\mathbf{\pm{0.04}}}$ & $1.57^{\pm{0.02}}$  \\ \bottomrule
  \end{tabular}
  }
  \caption{Quantitative comparison on Inter-X-e dataset for unconditional human reaction generation task.}
  \label{tab:unconditional Quantitative comparison}
\end{table}

\section{Ablation Study on Loss Items}
\label{supsec: Ablation Study on Loss Items}
The training loss of the diffusion model is defined in Eq.~14 in the main text. We propose two novel loss terms, $L_{react}$ and $L_{emo}$, which specifically promote accurate interaction and emotion prediction, respectively. The other loss terms in Eq.~14 are standard components commonly used in previous studies. The ablation results presented in Table~\ref{tab: ablation on losses} demonstrate the effectiveness of our proposed loss terms.

\begin{table}
\centering
  \begin{tabular}{@{}lccc@{}}
    \toprule
    Method & FID $\downarrow$ & ACC $\uparrow$ \\ \midrule
    GT &$0.79^{\pm{0.03}}$ &$0.89$ \\ 
    w/o $L_{react}$ & $2.72^{\pm{0.01}}$ & $0.82$ \\
    w/o $L_{emo}$ & $2.37^{\pm{0.02}}$ & $0.72$ \\
    \textbf{ALL} & $\mathbf{1.94}^{\mathbf{\pm{0.03}}}$ & $\mathbf{0.86}$  \\ \bottomrule
  \end{tabular}
  \caption{Ablation results of our proposed loss terms.}
  \label{tab: ablation on losses}
\end{table}

\section{Effectiveness of Emotion Conditioning}
\label{supsec: Effectiveness of Emotion Conditioning}

\begin{figure}
  \centering
  \includegraphics[width=1.0\linewidth]{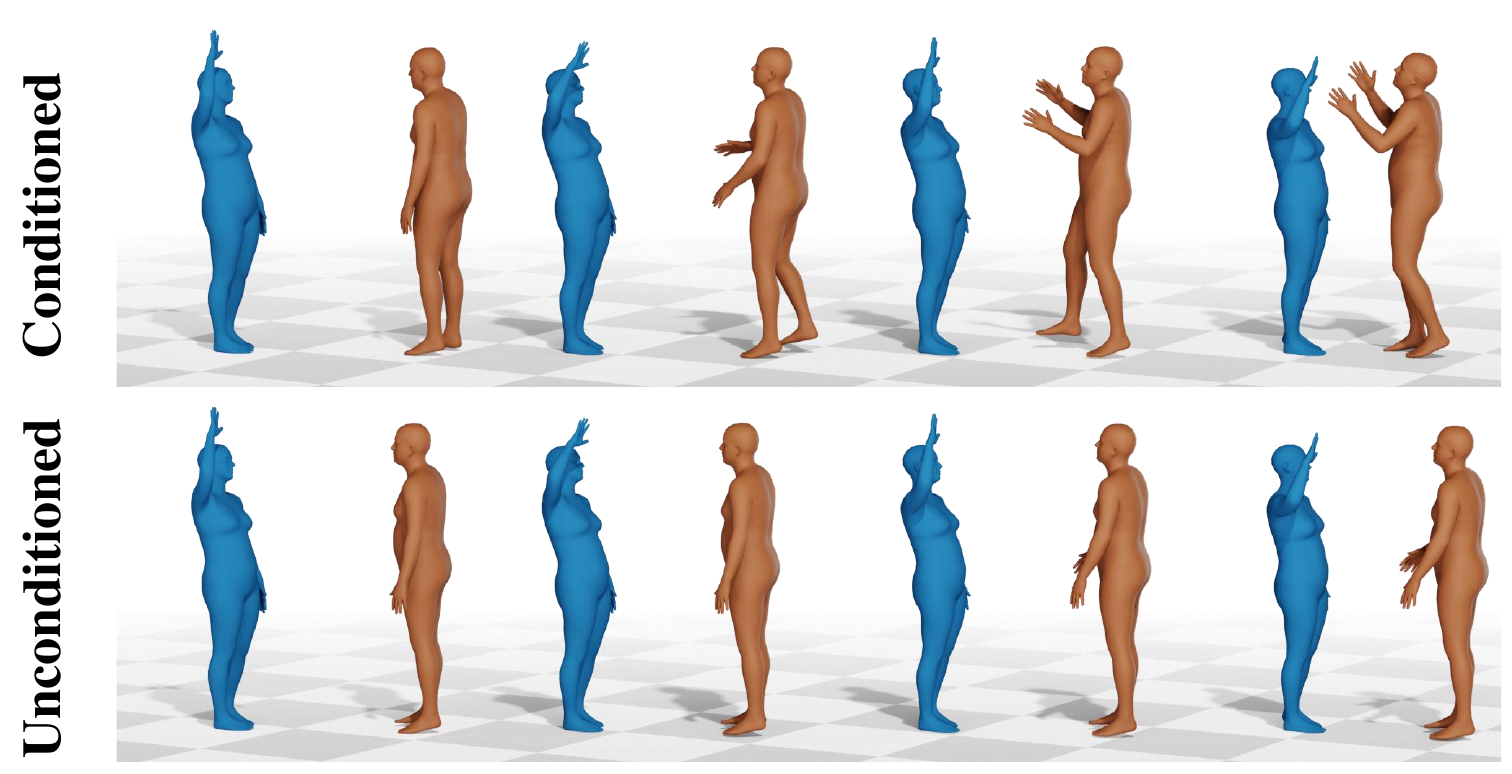}
  \caption{Visualization of Emotion Conditioning. The actor~(blue) is enthusiastically waving to greet. The top sequence represents reaction generation guided by emotion, while the bottom is generated without emotional conditions.}
  \label{fig:ab1}
\end{figure}

To further demonstrate the effect of emotion-driven generation, we compared the generated results under both emotion-conditioned and non-emotion-conditioned settings. In Figure~\ref{fig:ab1}, the responses from unconditional generation appear rigid. In comparison, the emotion-guided reaction produces results that align with the given emotion, while also exhibiting a clearer sense of interaction awareness. 

\begin{figure}
  \centering
  \includegraphics[width=1.0\linewidth]{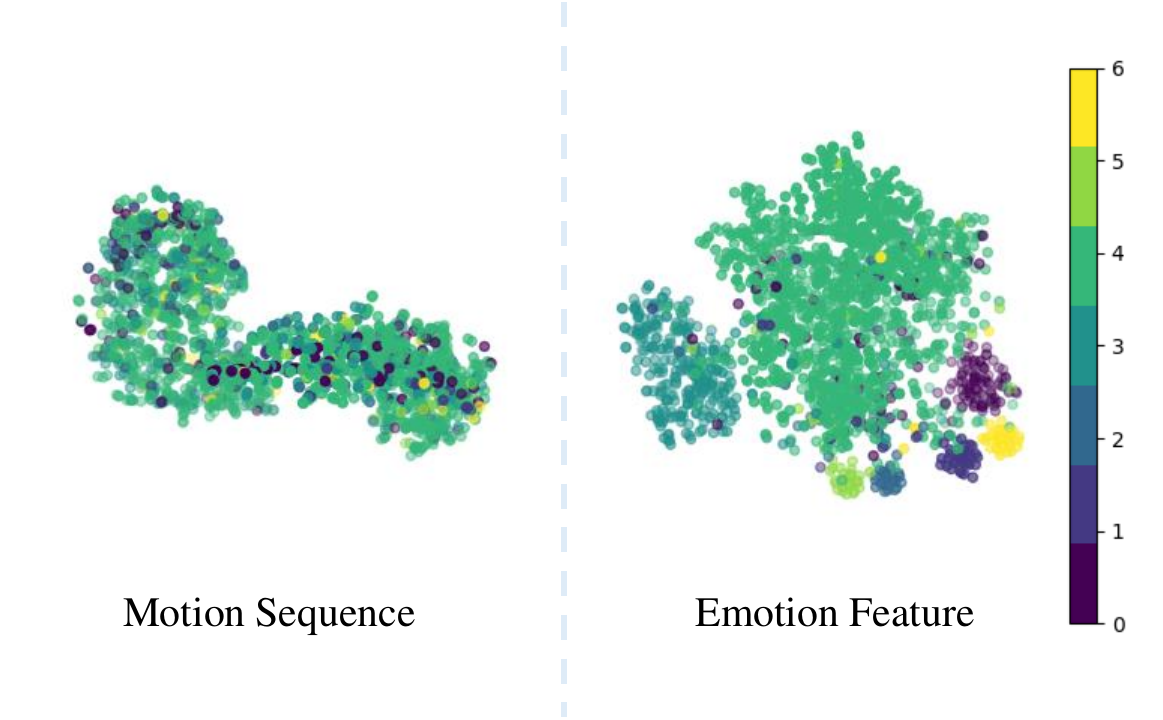}
  \caption{T-SNE visualization of distribution of origin motion and emotion feature.}
  \label{fig:tsne}
\end{figure}

Furthermore, to assess the superiority of mapping emotions to a continuous latent distribution space, we compare the emotional features by emotion prior encoder with the original motion data. The visualization of t-SNE results in Figure~\ref{fig:tsne} shows that our affective encoder can successfully decouple emotional features from body posture and actions.

\section{Implementation details}
\label{supsec: Implementation details}
We trained the emotion prediction network in a semi-supervised manner. For sequences without emotion annotations, two motion clips were randomly extracted from each sequence to construct a pair. During training, we enforce the predictions of these two samples to be identical, as a short motion sequence is expected to exhibit the same emotion. The network consists of 8 transformer layers for extracting emotion features from the motion sequences, with each layer comprising six attention heads and a dropout rate of 0.3. The dimension of the latent space was set to 1024. We followed the interaction sequence in Inter-X dataset to define the active and passive parties, and the division of the dataset adhered to its inherent setup. The training took 23 hours on a piece of RTX 3090 GPU for 500 epochs.

In our actor-reactor diffusion model, we utilize the same number of transformer layers and latent dimensions to construct the latent motion space. However, each layer is adapted to incorporate eight attention heads, with a reduced dropout rate of 0.1. During inference, we employ the DDIM strategy~\cite{song2022denoisingdiffusionimplicitmodels} with 50 timesteps to accelerate the generation of reactions.

Our symmetrical network holistically considers actor-reactor spatial relationships alongside global emotional features. As demonstrated in the inference time measurements in Table~\ref{tab:inference time}, our model produces realistic and diverse reactions while also achieving runtime efficiency that surpasses other similar frameworks.

\begin{table}
\centering
\resizebox{\linewidth}{!}{
  \begin{tabular}{@{}l|ccccccc@{}}
    \toprule
    Methods & ReGenNet	& InterGen & Ours \\ \midrule
    Running Time (s) &	$6.642385$ &	$6.605281$	& $6.543127$ \\ \bottomrule
  \end{tabular}
  }
  \caption{Average inference time for emotion-conditioned human reaction generation task.}
  \label{tab:inference time}
  
\end{table}

\section{Emotional User Study}

\begin{figure*}[!ht]
  \centering
  \includegraphics[width=1.0\textwidth]{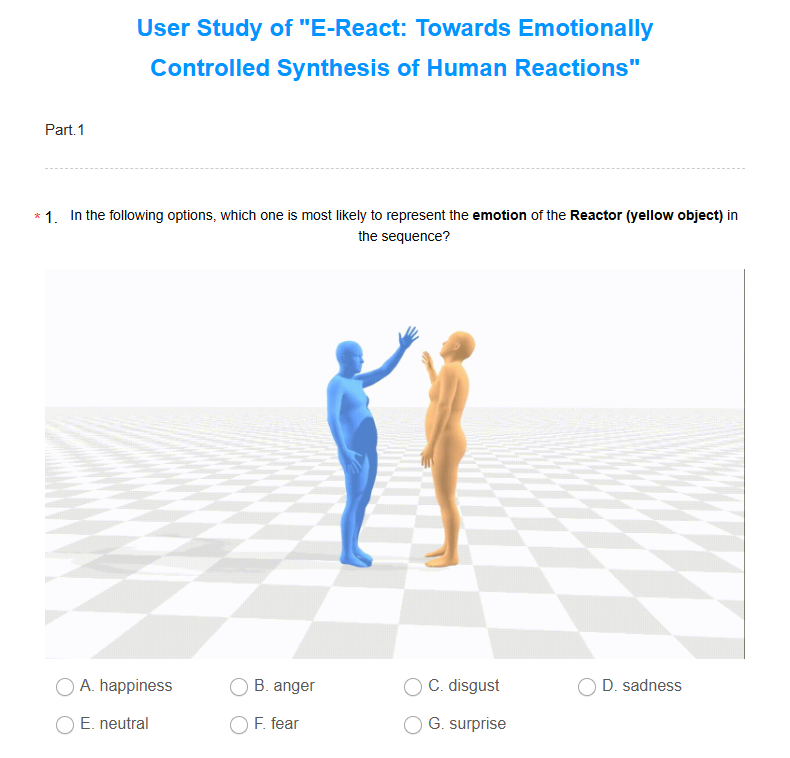}
  \vspace{-4mm}
  \caption{An example of a question from the part.1 in our survey.}
  \label{fig:userstudy_p1}
\end{figure*}

Following~\cite{carrasco2024level}, we developed a survey questionnaire to compare the consistency of our model's emotional judgments with those of humans. The questionnaire comprised 25 reactions generated under specified emotions and 25 reactions that were the ground true~(Part.1, as shown in Figure~\ref{fig:userstudy_p1}). A voting mechanism is employed to determine the emotion assigned to each sequence, with the emotion receiving the highest score representing the group consensus.

We compute the accuracy, recall, and F1 score based on the emotions identified by participants, ground-truth annotations, and model predictions.

\begin{equation} 
    Accuracy = 
    \frac{TP+TN}{TP+TN +FP+FN},
\end{equation}

\begin{equation} 
    Recall = 
    \frac{TP}{TP+FP},
\end{equation}
where $TP$ and $TN$ represent the probabilities of correctly predicting positive samples as positive and negative samples as negative, respectively. $FP$ and $FN$ denote the probabilities of incorrectly predicting positive samples as negative and negative samples as positive, respectively.

Additionally, we utilize the Jaccard index to analyze the consistency of emotion judgments. The results in Table~\ref{tab:user study} demonstrate strong agreement between user evaluations and our annotated emotions. Furthermore, the consistency between user evaluations and annotated emotions is comparable to that between user evaluations and controlled conditions, indicating that our model produces reactions closely aligned with human judgment.

\begin{figure*}[!ht]
  \centering
  \includegraphics[width=1.0\textwidth]{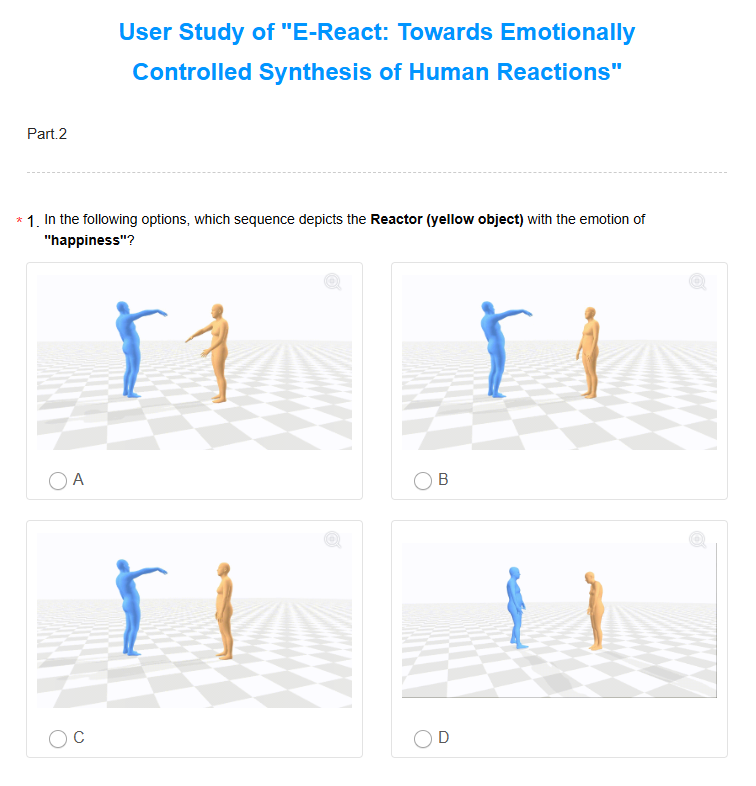}
  \vspace{-4mm}
  \caption{An example of a question from the part.2 in our survey.}
  \label{fig:userstudy_p2}
\end{figure*}

To evaluate the perceptual quality of emotion-driven reaction generation, we conducted a comparative user study~(Part.2, as shown in Figure~\ref{fig:userstudy_p2}) through carefully designed questionnaires, collecting human preferences between generated reactions from various models and ground-truth reference motions. We provide four reactions from Ground Truth, our method, InterGen, and ReGenNet for a given actor's motion and emotion. The same participants are asked to identify which reaction is the best. Each questionnaire includes 5 different interactive sequences. When a method's reaction is selected, it is counted as one score. The results shown in Table~\ref{tab:userstudy2} demonstrate that our method can produce more realistic reactive motions than the other methods.

\begin{table}
\centering
\resizebox{\linewidth}{!}{
  \begin{tabular}{@{}ccccc@{}}
    \toprule
    Motions & Precision $\uparrow$  & Recall $\uparrow$  & F1-Score $\uparrow$  & Jaccard's index $\uparrow$ \\ \midrule
    GT &$0.73$ &$0.65$ &$0.68$ &$0.3713$ \\ 
    Generated &$0.68$ &$0.59$ &$0.63$ &$0.3472$ \\ \bottomrule
  \end{tabular}
  }
  \caption{User study Part.2: Results demonstrating consistency between our model predictions and human emotional judgments.}
  \label{tab:user study}
\end{table}

\begin{table}
\centering
\resizebox{\linewidth}{!}{
\setlength{\tabcolsep}{0.05\columnwidth}
  \begin{tabular}{@{}l|c|ccc}
    \toprule
    Method & GT & ReGenNet & InterGen & \textbf{Ours}\\ \midrule
    Score $\uparrow$ & 87 & 34 & 46 & \textbf{83} \\
\bottomrule
  \end{tabular}
 }
  \caption{User study Part.1: The number of times each method's reaction is selected.}
\label{tab:userstudy2}

\end{table}

\section{More Visual Results of Applications}
\begin{figure*}[!ht]
  \centering
  \includegraphics[width=1.0\textwidth]{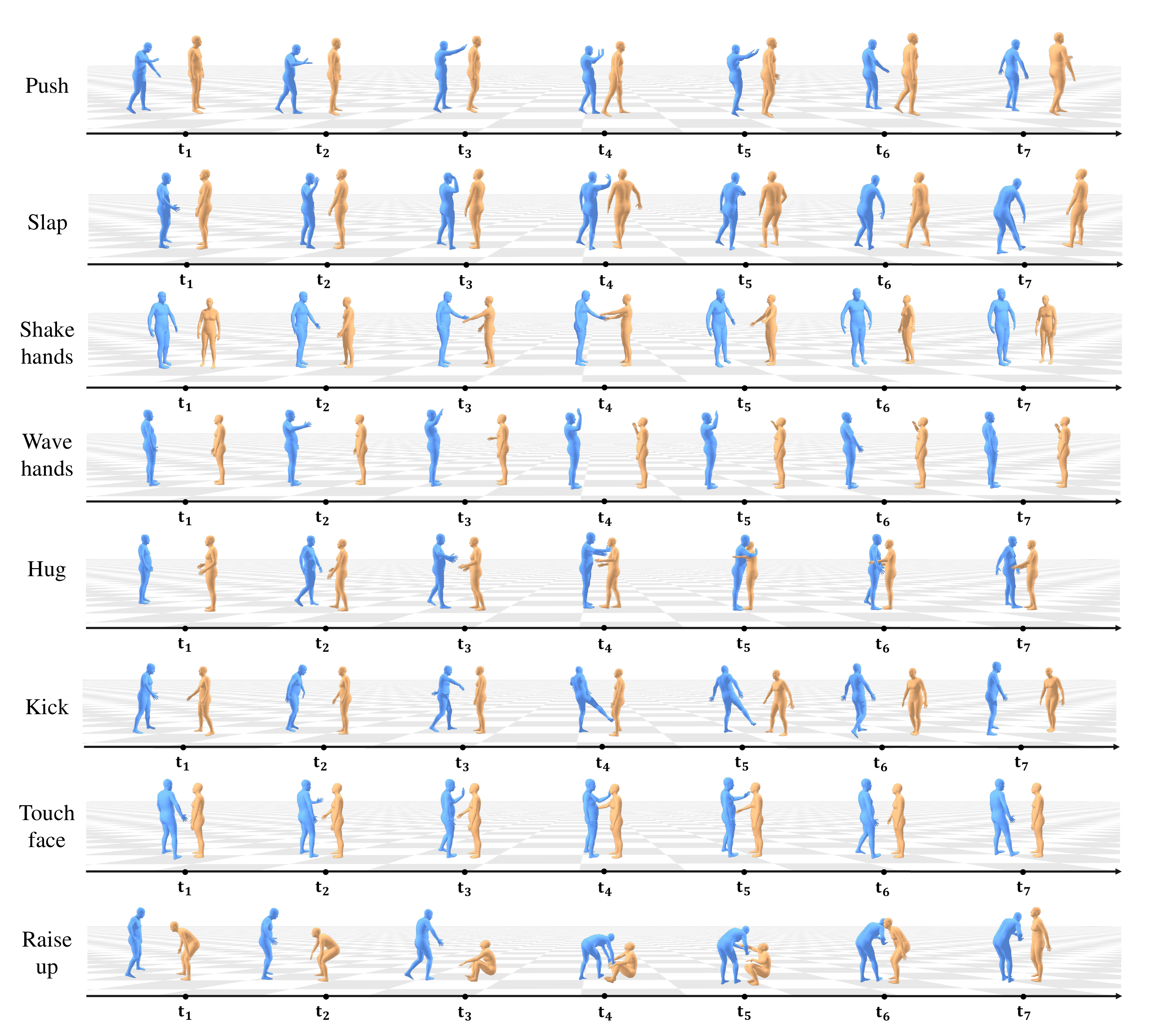}
  \vspace{-4mm}
  \caption{Visualizations of generated reactions utilizing our method.}
  \label{fig:viz_results}
\end{figure*}

We further discuss the application content of Emotion-driven Reaction Generation in the main text. For \textbf{Empathetic Reaction Generation}, we first estimate the emotion of the action using an emotion prediction network before the reaction generation process. This estimated emotion is then used as an emotional condition to drive the reactor's reaction generation. Under this setting, the reactor can automatically perceive the actor's emotional state and generate a synchronized reaction. We have outlined the results of Empathetic Reaction Generation in common interactive scenarios encountered in daily life, as shown in Figure~\ref{fig:viz_results}. It can be observed that empathetic reactions, as a setting close to real-world dynamics, facilitate the generation of natural and vivid interactive movements.
For \textbf{Emotional Editing of Reactions}, qualitative comparison experiments have demonstrated that our model is capable of generating different reactions by modifying the controlled emotions, which related work cannot achieve. We showcase more examples of emotional editing in the video provided in the supplementary demo.

\end{document}